\documentclass[sigconf,nonacm]{acmart}

\AtBeginDocument{%
  }

\setcopyright{none}
\copyrightyear{2026}
\acmYear{2026}

\usepackage{tikz}
\usepackage{pgfplots}
\pgfplotsset{compat=1.18}
\usetikzlibrary{positioning,arrows.meta,shapes.geometric,fit,backgrounds,calc}
\usepackage{booktabs}

\begin{document}

\title{Agentic ML Exploration (A-MLE) for Ads Ranking}

% Author block --- update / re-order as needed before submission.
\author{
    Erwin Gao,
    Vinodh Kumar Sunkara,
    Jingyi Guan,
    Qinjin Jia, 
    Hangjun Xu,
    Xiang Ji,
    Sherman Wong,
    Surya Teja Chavali,
    Pratik Vaishnavi,
    Aryan Pandhi,
    Xiaoyu Deng,
    Zhaodong Wang,
    Samarth Inani,
    Fan Yang,
    Jakob Moberg,
    Zoe Zu,
    Nicolas Bievre,
    Sami Khenissi,
    Amit Jaspal,
    Ehsan Fakharizadi,
    Srinidhi Viswanathan,
    Dorothy Sun,
    Abishek Vanam,
    Sneha Iyer,
    Sheela Yadawad,
    Wenjie Chen,
    Gaby Nahum,
    Junhua Gu,
    Peter Chu,
    Yucheng Liu,
    Xin Zhao,
    Vitor Cid,
    Chaorong Chen,
    Vijay Pappu,
    Ashwin Kumar,
    Wenlin Chen,
    Ben Schulte, 
    Deepak Chandra,
    Ritwik Tewari
}

\affiliation{
  \institution{Meta Platforms, Inc.}
  \city{Menlo Park and New York}
  \country{USA}
}

\email{clingsz, vinodhsunkara, jingyiguan, qjia, hangjunxu, xiangji, shermanwong, teja5832, pratikv, aryanpandhi (@meta.com)}
\email{xiaoyud, zhaodongwang, samarthinani, fyang7, jamoberg, zoezu, nbievre, samikhenissi, ajaspal (@meta.com)}
\email{ehsanf, srinidhiv, dorothysun, abishekvanam, snehaiyer, sheelayadawad, wenjiec, gnahum12345, jug, pchu (@meta.com)}
\email{yuchengliu, zhaox, cid, chaorong, psnvijay, ashwink3029, wenlinchen, bschulte, deepakchandra,  ritwikt (@meta.com)}

%% ---------------------------------------------------------------------
%% Abstract
%% ---------------------------------------------------------------------
\begin{abstract}
Modern industrial ads ranking stacks are increasingly bottlenecked not by model capacity or training compute, but by the throughput of \emph{human ML iteration} -- the cycles of research, implementation, training, debugging, evaluation, and launch required to surface a single statistically significant improvement. A typical ranking stack contains numerous differentiated models with heterogeneous data, architectures, and infrastructure constraints, and each cycle takes days to weeks of senior engineer attention per model. As a result, techniques that have proven effective on one model diffuse into others slowly and unevenly, leaving substantial recoverable signal unexplored.

We present \textbf{Agentic ML Exploration (A-MLE)}, an autonomous LLM-agent system that systematically explores ML techniques across a portfolio of ads ranking models. A-MLE decomposes ML iteration into five stages involving hypothesis generation, exploration strategy, experiment execution, result analysis and shared knowledge substrate which are orchestrated by a single agent that invokes domain-specific skills and agentic workflows against a sandboxed execution layer, with human-in-the-loop checkpoints at each stage boundary.

We deploy A-MLE across a representative set of large-scale ads ranking models and evaluate it along a tiered capability framework (tool availability, autonomous workflow execution, and open-ended exploration). We further report a controlled cross-LLM study using a fixed agent loop, which surfaces qualitative differences in execution reliability and exploration aggressiveness across the Claude Sonnet, Gemini, and GPT families.

We discuss failure modes and the design choices that govern reliability. Our findings suggest that agentic exploration is a practical force multiplier for ML engineers in industrial recommenders, especially for the long tail of models that rarely receive expert attention.
\end{abstract}

%% CCS



\maketitle

%% ---------------------------------------------------------------------
%% Introduction
%% ---------------------------------------------------------------------
\section{Introduction} \label{sec:introduction}
Industrial ads ranking systems have grown into portfolios of dozens of differentiated models, each tailored to a specific objective (e.g., click, conversion, view), surface, and ad segment. Each model carries its own training data, feature set, architecture, and infrastructure constraints, which makes the effort of porting a proven idea from one model to another surprisingly large.

The \emph{rate of progress} on such a portfolio is therefore governed less by the ceiling of any single architectural innovation and more by the \emph{throughput} of human ML iteration. A single end-to-end exploration from hypothesis generation and exploration code changes to successful model training and proposal takes a senior ML engineer on the order of days to weeks per model. With a finite engineering pool, only a small subset of model $\times$ technique combinations are ever attempted, and the long tail of models receives little exploration even when proven techniques exist nearby.

Recent advances in large language model (LLM) agents \cite{yao2023react,wang2024voyager,park2023generative,xi2023rise} suggest a different operating point: rather than improving any one model architecture, we can improve the \emph{iteration loop itself}. We design and study \textbf{A-MLE}, a system in which an LLM agent that is equipped with a domain-specific skill library, a code execution environment, and structured access to training and evaluation pipelines - autonomously explores ML techniques across a portfolio of ranking models. The agent handles hypothesis generation, exploration planning, experiment execution, result analysis, etc. stages in the ML Exploration cycle with human engineers serving as reviewers at well-defined checkpoints rather than as moment-to-moment operators.

\paragraph{Contributions.}
\begin{itemize}
  \item We characterize the manual-iteration bottleneck in industrial ads ranking and frame it as the relevant unit of system-level optimization (Section~\ref{sec:problem}).
  \item We describe the A-MLE architecture as a single agent orchestrating five stages over a shared skill library, sandbox, and the design choices that govern its reliability (Section~\ref{sec:system}).
  \item We deploy A-MLE across a representative model portfolio and report results along a tiered capability framework, including a concrete model-improvement headline and a controlled cross-LLM comparison (Sections~\ref{sec:setup} and \ref{sec:results}).
  \item We document the dominant failure modes, orchestration harness that, in our experience, dominate base-model capability (Section~\ref{sec:results}).
\end{itemize}

The remainder of the paper is organized as follows. Section~\ref{sec:related} reviews related work in AutoML and LLM agents. Section~\ref{sec:problem} formalizes the manual-iteration bottleneck. Section~\ref{sec:system} presents the A-MLE system. Sections~\ref{sec:setup} and \ref{sec:results} cover experimental setup and results. Section~\ref{sec:conclusion} concludes.

%% ---------------------------------------------------------------------
%% Related Work
%% ---------------------------------------------------------------------
\section{Related Work} \label{sec:related}

\paragraph{AutoML and Neural Architecture Search.} A long line of work has automated parts of the ML pipeline including hyperparameter optimization \cite{snoek2012practical,bergstra2012random}, neural architecture search \cite{zoph2017nas,liu2019darts,real2019regularized}, and end-to-end AutoML systems \cite{feurer2015autosklearn,jin2019autokeras}. These systems target a well-defined search space and a single optimization objective. A-MLE differs in that the search space is open-ended (any code-level change to a complex model architecture) and the objective is composite (offline metric gain, infrastructure feasibility, statistical significance, and launch candidate quality).

\paragraph{LLM Agents and Tool Use.} Recent work has shown that LLMs equipped with tools and feedback can solve increasingly complex multi-step tasks \cite{yao2023react,schick2023toolformer,shinn2023reflexion,wang2024survey}. Code-generation agents \cite{chen2021codex,roziere2023codellama,jimenez2024swebench} have demonstrated competence on software engineering benchmarks. A-MLE extends this line into the ML engineering domain, where tasks are dominated not by code synthesis but by the surrounding loop of hypothesis generation, adaptive training under failures, evaluation, and identifying high-ROI candidates under infrastructure and compute constraints.

\paragraph{ML Engineering Automation.} Concurrent and prior work has explored agents for data science \cite{huang2024mlbench,hong2024data,guo2024dsbench} and for end-to-end ML research \cite{huang2025mlebench,lu2024aiscientist}. These efforts focus primarily on academic Kaggle-style benchmarks. A-MLE targets a complementary regime: numerous industry scale models, where each iteration carries significant compute cost, training datasets are large, and the deliverable is incremental improvements on top of a mature baseline with statistical rigor, beyond a leaderboard score as the gating criterion.

\paragraph{Recommendation and Ads Ranking.} Modern ads ranking systems use cascaded multi-stage architectures with deep models at each stage \cite{naumov2019dlrm,covington2016deep,zhou2018din,zhou2019dien,zhai2024actions}. Recent advances include self-supervised learning for sparse features \cite{yao2021ssl}, token-mixing architectures \cite{tolstikhin2021mlp,zhu2025rankmixerscalingrankingmodels}, many other embedding based features. A-MLE treats this body of techniques as a \emph{search space}: the agent's job is to identify which technique to attempt next on which model, given the current state of evidence.

%% ---------------------------------------------------------------------
%% Problem
%% ---------------------------------------------------------------------
\section{The Manual ML Iteration Bottleneck} \label{sec:problem}

A canonical manual iteration on a single ranking model consists of multiple phases including:
\begin{enumerate}
  \item \textbf{Ideation and hypothesis generation.} An engineer surveys recent literature, internal proposals, and the model's recent training history to select a candidate technique to try.
  \item \textbf{Prioritize exploration candidates.} With a given training compute budget to trigger the model runs, an effective prioritization is required to identify which of the initial set of hypotheses are worth pursuing and number of variants to attempt per hypothesis.
  \item \textbf{Implementation.} The engineer writes or adapts the code change within the model architecture and against its training entry point, taking care not to break unrelated downstream consumers.
  \item \textbf{Training and failure recovery.} The engineer launches a training run on a shared compute pool, monitors stability, and intervenes when the run fails.
  \item \textbf{Evaluation triage.} The engineer compares offline metrics against the rolling baseline, decomposes gains by different required ad and user segments, and decides whether the candidate is worth promoting.
  \item \textbf{Proposal preparation.} The engineer authors a proposal with experiment design, results, statistical analysis, and a launch recommendation.
\end{enumerate}

Each phase has different challenges. Ideation and hypothesis generation is bottlenecked by the engineer's familiarity with both the model and the broader technique landscape. Prioritization of candidates for model training is dependent on in-depth understanding of the trade offs with different directions of exploration. Implementation is bottlenecked by the complexity of the codebase and of the technique itself. Training and observation are bottlenecked by infrastructure stochasticity, the shared-pool preemption, data pipeline incidents, and intermittent evaluation failures. Evaluation triage is bottlenecked by metric variance and the engineer's ability to disentangle real signal from baseline drift.

The aggregate effect is that a single end-to-end iteration on a single model takes a multi-week span. Multiplied across a portfolio of models, the total number of (model, technique) pairs that can be explored within any given period is inherently limited. This is the gap that A-MLE targets.

%% ---------------------------------------------------------------------
%% System Design
%% ---------------------------------------------------------------------
\section{System Design} \label{sec:system}
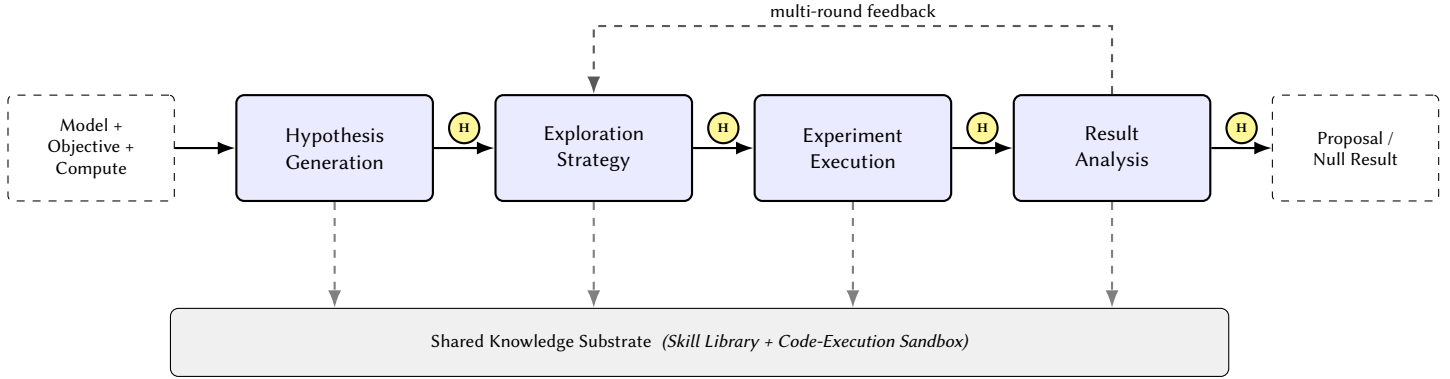
\begin{figure*}[t]
\centering
\begin{tikzpicture}[
  stage/.style   = {rectangle, draw, rounded corners=3pt, thick,
                    minimum height=14mm, minimum width=26mm,
                    align=center, fill=blue!8, font=\small\sffamily},
  io/.style      = {rectangle, draw, dashed, rounded corners=2pt,
                    minimum height=14mm, minimum width=22mm,
                    align=center, font=\footnotesize\sffamily},
  substrate/.style={rectangle, draw, rounded corners=3pt,
                    fill=gray!12, font=\footnotesize\sffamily,
                    minimum height=9mm, minimum width=140mm,
                    align=center},
  hitl/.style    = {circle, draw, fill=yellow!50, inner sep=0pt,
                    minimum size=4mm, font=\tiny\bfseries},
  flow/.style    = {-Latex, thick},
  feedback/.style= {-Latex, dashed, thick, gray!70!black},
]
\node[io]                          (in)  at (0,0)    {Model +\\ Objective +\\ Compute};
\node[stage, right=8mm of in]      (s1)              {Hypothesis\\Generation};
\node[stage, right=8mm of s1]      (s2)              {Exploration\\Strategy};
\node[stage, right=8mm of s2]      (s3)              {Experiment\\Execution};
\node[stage, right=8mm of s3]      (s4)              {Result\\Analysis};
\node[io,    right=8mm of s4]      (out)             {Proposal /\\ Null Result};
\draw[flow] (in) -- (s1);
\draw[flow] (s1) -- (s2) node[hitl, midway, above=0.5mm] {H};
\draw[flow] (s2) -- (s3) node[hitl, midway, above=0.5mm] {H};
\draw[flow] (s3) -- (s4) node[hitl, midway, above=0.5mm] {H};
\draw[flow] (s4) -- (out) node[hitl, midway, above=0.5mm] {H};
\draw[feedback]
  (s4.north) -- ++(0,9mm) -| (s2.north)
  node[pos=0.25, above, font=\footnotesize\sffamily, black]
       {multi-round feedback};
\node[substrate, below=14mm of s2.south, xshift=14mm] (sub)
      {Shared Knowledge Substrate \;\textit{(Skill Library + Code-Execution Sandbox)}};
\foreach \n in {s1,s2,s3,s4}{
  \draw[flow, dashed, gray] (\n.south) -- (\n.south |- sub.north);
}
\end{tikzpicture}
\caption{A-MLE orchestration. Within each session, a single agent walks five stages with human-in-the-loop checkpoints (H) and a multi-round feedback loop into strategy planning. Across sessions, the agent reads from and writes back to a shared substrate (per-technique and per-model artifacts) so that outcomes accumulated on one model become consumable on another.}
\label{fig:mlea_overview}
\end{figure*}

A-MLE is organized around five stages primarily -- \emph{hypothesis generation}, \emph{exploration strategy}, \emph{experiment execution}, \emph{result analysis}, \emph{shared substrate} -- driven by a single tool-using LLM agent that operates against a shared skill library and a sandboxed code-execution layer. Each \emph{exploration session} is parameterized by a (model, objective, compute) triple and produces, at termination, either a documented proposal or a documented null result. Figure~\ref{fig:mlea_overview} shows the overall flow; we describe each stage below.

\subsection{Hypothesis Generation}
The agent reads the model's recent training configuration, baseline metrics, and the rolling history of attempted techniques, then proposes a small set of candidate techniques, each with an explicit rationale linking the technique to the model's current state. Hypotheses can be sourced from internal generators e.g. model-internal-state analyzers, training-efficiency analyzers, recent-literature retrievers, and are scored by an LLM critic for novelty and feasibility. Grounding hypotheses in the live state of the model rather than a stale snapshot is the focus here: a hypothesis that fits the previous baseline rarely transfers cleanly after a baseline refresh.

\subsection{Exploration Strategy}
Given different constraints (number of training runs, total compute, or wall time) and a candidate set, the agent prioritizes and plans the experiment model iterations sequence. Plans typically interleave \emph{exploration} -- validating individual hypotheses in isolation -- with \emph{exploitation} -- combining the most promising candidates and pushing them harder. The agent negotiates the plan with the user at this checkpoint, since trade-offs between aggressiveness and compute are best surfaced explicitly to review.

\subsection{Experiment Execution}
The agent edits the training configuration or the model architecture in a sandboxed copy of the codebase, runs type checks and unit tests, builds an image with the updated code, runs a short smoke pass for verification, then submits the full training job. Once submitted, the agent monitors progress, distinguishes infrastructure errors from genuine training divergence, and either retries, fixes at the code or config level, or reports the run as needed. Across a multi-experiment plan, the agent adapts through failures e.g. if one branch is found to be infeasible, it reroutes the remaining compute rather than abandoning the session.

\subsection{Result Analysis}
After each completed run, the agent computes statistical significance against the rolling baseline (not a frozen snapshot, which would be vulnerable to baseline drift), decomposes metrics by segment to surface localized regressions, and triggers an automatic re-run when within-run variance exceeds a threshold. At the end of an exploration round the agent produces a structured leaderboard of candidates and either feeds it back into the strategy stage for another round or assembles a final proposal.

\subsection{Shared Substrate}
Each session reads from and writes to a shared-learning substrate: a structured, machine-actionable representation of cross-model ML knowledge. Knowledge accumulated on one model becomes automatically consumable on another model. The substrate is long-lived markdown trees versioned in source control with per-technique and per-model shared knowledge from the past and applied on subsequent iterations. At session start, the agent matches the target model's context against the substrate's structured eligibility annotations to surface ML techniques with prior evidence on architecturally similar models. At session end, the new outcomes are then committed back to the relevant Track Record,  reviewable like any other source-control change.

\subsection{Human-in-the-Loop Checkpoints}
In addition to the five stages, despite the agent's autonomy within each stage, every stage boundary is a checkpoint at which a human engineer can approve continuation, request modifications, or terminate the session. The agent's strength is in covering a wide search space efficiently; the engineer's strength is in reviewing the findings to cover all edge cases and gaps, informed by practical expertise in trade-offs. The checkpoint structure lets us combine the two without giving up either, and limits the blast radius of any single agent decision e.g. a hallucinated code change is caught before training launch, a miscalibrated evaluation is caught before proposal authoring.

%% ---------------------------------------------------------------------
%% Experimental Setup
%% ---------------------------------------------------------------------
\section{Experimental Setup} \label{sec:setup}

\paragraph{Models.} We deploy A-MLE against a representative set of large-scale ads ranking models drawn from a real industrial portfolio. The set is chosen to span the principal axes of variation: optimization objective (click vs.\ conversion vs.\ view), surface, and architecture family (deep cross network, deep interest network, multi-tower). For external review purposes, models are referred to as anonymized identifiers $M_1, M_2, \dots$ that abstract away surface and product-line specifics. Several detailed studies in the remainder of the paper are conducted on a single experimental ranking model that we denote $M^\ast$ -- a regression-objective model with a lightweight resource footprint, suitable as a benchmark substrate.

\paragraph{Tiered capability framework.} To make the setup tractable for evaluation, we organize tasks in a three-tier capability framework, drawn from our internal benchmark design:
\begin{itemize}
  \item \textbf{L1 -- Tool availability.} The agent is asked focused single-step questions over a model's training configuration, evaluation strategy, metric stores, serving and training infrastructure etc. This isolates the question of whether the agent has the right APIs and domain knowledge.
  \item \textbf{L2 -- Autonomous workflow execution.} The agent is given a multi-step task that requires submitting workflows, monitoring them, recovering from infrastructure failures, and summarizing results. This isolates the question of whether the agent can reliably operate the iteration loop.
  \item \textbf{L3 -- Open-ended exploration.} The agent is given a model, an objective, and compute, and asked to surface the best improvement it can find. This is the regime we ultimately care about.
\end{itemize}

\paragraph{Baselines.} We use \emph{rolling baseline} to refer to the current reference training configuration of a given model — the metric values against which candidate changes are compared. Additionally, we also use \emph{methodological baselines} (manual and semi-automated) which are operating points we compare A-MLE against. The \textbf{manual} baseline is the prior operating point: senior ML engineers executing iterations themselves. The \textbf{semi-automated} baseline retains the engineer in the driver's seat for hypothesis generation but uses scripted helpers for triggering runs, evaluation, etc. steps. Both baselines share the same offline evaluation suite as A-MLE.

\paragraph{Metrics.} We report five classes of metrics:
\begin{itemize}
  \item \textbf{Throughput.} Completed end-to-end iterations per engineer-week, where a completed iteration terminates in either a documented proposal or a documented null result.
  \item \textbf{Training success rate.} Fraction of training runs the agent triggers that complete successfully (after automated debugging and retries) without requiring human intervention.
  \item \textbf{Proposal acceptance rate.} Fraction of agent-authored proposals with statistically significant offline impact that pass human review gating without rework.
  \item \textbf{Technique coverage.} Distinct technique families surfaced across the model portfolio over a fixed evaluation window.
  \item \textbf{Model Offline evaluation.} Model iterations evaluated using offline metric suite like NE (Normalized Entropy) NE for prediction quality and informativeness, rMSE for calibration between predicted and observed rates etc.
\end{itemize}

\paragraph{Reporting.} All performance results are reported as \emph{relative} improvements over the relevant baseline. Statistical significance is measured and reported against the rolling-baseline framework described in Section~\ref{sec:system}.

%% ---------------------------------------------------------------------
%% Results
%% ---------------------------------------------------------------------
\section{Results, Analysis, and Discussion} \label{sec:results}

\subsection{End-to-End Throughput} \label{subsec:throughput}
Across the evaluation window, A-MLE delivered multiple times the productivity in completed iterations per engineer-week compared to the baseline described in Section~\ref{sec:setup}. The semi-automated baseline showed a smaller order improvement in throughput with less stronger offline impact and longer iteration cycle to land convergence, indicating that the principal source of leverage is not the automation of any individual phase but the agent's ability to chain phases without engineer-mediated handoffs.

\subsection{Training Success Rate}
The fraction of A-MLE-triggered runs that completed successfully, after the agent's automated debugging and bounded re-run loop -- meaningfully surpassed the baseline described in Section~\ref{sec:setup}. A small minority of runs still required human attention when the agent failed to recover within its retry limits set, which we cap to ensure efficient use of training resources.

\subsection{Proposal Acceptance Rate}
A-MLE-authored proposals passed review criteria at a much higher rate than that of the baseline described in Section~\ref{sec:setup}. Reviewers cited better-grounded statistical analysis, cleaner documentation of negative results, and explicit segment-level decomposition as the differentiating factors.

\subsection{Tier-1 Capability Bench} \label{subsec:l1}

\begin{figure}[t]
\centering
\begin{tikzpicture}
  \begin{axis}[
    title={\sffamily\small L1 capability bench (\% correct)},
    width=8.6cm, height=5.6cm,
    ybar=2pt,
    bar width=8pt,
    enlarge x limits=0.18,
    ymin=0, ymax=110,
    ytick={0,25,50,75,100},
    ylabel={\footnotesize \% correct},
    symbolic x coords={Result Analysis, Job Modification, Efficiency, Overall},
    xtick=data,
    x tick label style={font=\scriptsize},
    legend style={
      at={(0.5,-0.20)}, anchor=north,
      legend columns=3,
      font=\scriptsize, draw=none, fill=none,
      /tikz/every even column/.append style={column sep=6pt},
    },
    nodes near coords,
    nodes near coords style={font=\tiny},
    grid=major,
    grid style={dashed, gray!25},
  ]
    \addplot+[fill=gray!50, draw=gray!70!black] coordinates {
      (Result Analysis,12.5) (Job Modification,9) (Efficiency,0) (Overall,8)
    };
    \addplot+[fill=blue!35, draw=blue!60!black] coordinates {
      (Result Analysis,12.5) (Job Modification,36) (Efficiency,0) (Overall,16)
    };
    \addplot+[fill=orange!70, draw=orange!80!black] coordinates {
      (Result Analysis,37.5) (Job Modification,100) (Efficiency,50) (Overall,68)
    };
    \legend{Generic LLM, ML agent (generic), ML agent (domain-equipped)}
  \end{axis}
\end{tikzpicture}
\caption{L1 tool-availability bench on $M^\ast$. The domain-equipped configuration is necessary to clear the basic operations bar; generic LLM capability is not sufficient.}
\label{fig:l1bench}
\end{figure}
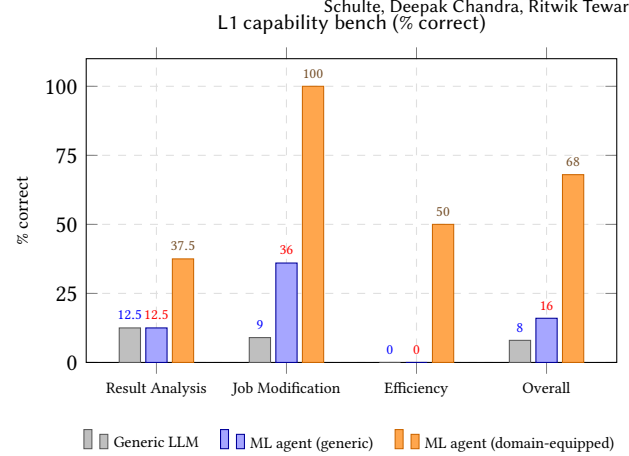

Figure~\ref{fig:l1bench} reports the L1 capability bench on $M^\ast$, comparing three configurations: a generic LLM with no ML tooling, a generic ML agent with cross-portfolio tools but no domain knowledge, and the domain-equipped A-MLE configuration. The domain-equipped agent reaches $68\%$ overall accuracy versus $16\%$ for the generic ML agent and $8\%$ for the generic LLM. The largest gap is on \emph{job-config modification}, where the domain-equipped agent solves all questions; the generic configurations score below $40\%$. This confirms that domain skills, not raw LLM capability, dominate at the tool-availability tier.

\subsection{Tier-2 Workflow Execution} \label{subsec:l2}
At the workflow-execution tier, we evaluate the agent on multi-step tasks that require submitting training workflows, monitoring asynchronous completions, recovering from infrastructure failures, and summarizing results. Four tier-2 representative tasks anchor this bench on $M^\ast$: (i) a \emph{baseline refresh}, in which the agent refreshes a baseline workflow onto the latest available training window and reports train/eval metrics; (ii) a \emph{variance test}, in which the agent submits multiple duplicate runs of the same baseline and computes train/eval/QPS variance across them; (iii) a \emph{config-change experiment}, in which the agent clones a baseline, applies a structured architectural edit (e.g., doubling a layer width), runs both versions to completion, and produces a side-by-side metric comparison; and (iv) a \emph{batch offline evaluation}, in which the agent schedules offline-evaluation runs across a set of previously trained jobs and aggregates the resulting metric rollups for side-by-side comparison.

The domain-equipped A-MLE configuration completes all four tier-2 tasks end-to-end with high reliability. Three capabilities differentiate it from a single-turn LLM loop: (a) an explicit \emph{waiting operator} that lets the agent suspend its own loop pending an external event (workflow completion, eval availability) and resume cleanly when the condition is met, which is necessary to support multi-hour asynchronous training jobs; (b) the ability to distinguish infrastructure failures from genuine training divergence and retry without operator help; and (c) reliable summarization that survives the noise of multi-run, multi-stage output. We defer cross-LLM variation on these tasks to Section~\ref{subsec:llm}.

\subsection{Tier-3 Exploration Outcomes} \label{subsec:l3}
At the open-ended exploration tier, A-MLE delivered measurable offline improvements on a majority of evaluated models. Table~\ref{tab:improvements} summarizes representative headline results on $M^\ast$ across two generations of the system: a single-hypothesis arch scale-up exploration and a multi-source exploration that combines hypothesis generators (model-internal-state analyzers and training-efficiency analyzers). The multi-source variant produces a $+2.557\%$ relative improvement on the regression objective with neutral training throughput ($+0.42\%$ QPS), substantially above the single-hypothesis baseline.

\begin{table}[t]
\centering
\small
\begin{tabular}{lcc}
\toprule
\textbf{A-MLE configuration} & \textbf{Rel.\ improvement} & \textbf{Train QPS impact} \\
\midrule
Single-hypothesis arch scale-up   & $+0.44\%$  & neutral  \\
Multi-round arch exploration       & $+0.58\%$  & neutral  \\
Multi-source (arch + efficiency)   & $+2.56\%$  & $+0.42\%$ \\
\bottomrule
\end{tabular}
\caption{Headline exploration outcomes on the experimental ranking model $M^\ast$. Improvements are relative offline regression-error reductions; QPS impact reports change in training throughput.}
\label{tab:improvements}
\end{table}

\subsection{Cross-LLM Comparison} \label{subsec:llm}
\begin{figure}[t]
\centering
\begin{tikzpicture}
  \begin{axis}[
    name=ax1,
    title={\sffamily\small (a) L2: workflow execution (avg.\ task completeness)},
    width=8.6cm, height=4.8cm,
    ybar,
    bar width=10pt,
    enlarge x limits=0.10,
    ymin=0, ymax=110,
    ytick={0,25,50,75,100},
    ylabel={\footnotesize Score (0--100)},
    symbolic x coords={Sonnet 3.5,Sonnet 3.7,Sonnet 4.0,Gemini 2.5,GPT-4,GPT-5},
    xtick=data,
    x tick label style={font=\scriptsize, rotate=30, anchor=east},
    nodes near coords,
    nodes near coords style={font=\scriptsize},
    grid=major,
    grid style={dashed, gray!25},
  ]
    \addplot+[fill=blue!55, draw=blue!70!black] coordinates {
      (Sonnet 3.5,93.3) (Sonnet 3.7,93.3) (Sonnet 4.0,96.7)
      (Gemini 2.5,100.0) (GPT-4,13.3) (GPT-5,66.7)
    };
  \end{axis}
  \begin{axis}[
    name=ax2,
    at={(ax1.below south west)},
    anchor=above north west,
    yshift=-8mm,
    title={\sffamily\small (b) L3: arch exploration outcome},
    width=8.6cm, height=4.8cm,
    ybar=2pt,
    bar width=7pt,
    enlarge x limits=0.14,
    ymin=0, ymax=290,
    ylabel={\footnotesize rMSE improvement ($\times 10^{-4}$)},
    symbolic x coords={Sonnet 3.5,Sonnet 3.7,Sonnet 4.0,Gemini 2.5,GPT-5},
    xtick=data,
    x tick label style={font=\scriptsize, rotate=20, anchor=east},
    nodes near coords,
    nodes near coords style={font=\tiny},
    legend style={at={(0.5,-0.32)}, anchor=north, legend columns=2,
                  font=\scriptsize, draw=none, fill=none,
                  /tikz/every even column/.append style={column sep=8pt}},
    grid=major,
    grid style={dashed, gray!25},
  ]
    \addplot+[fill=teal!55, draw=teal!70!black] coordinates {
      (Sonnet 3.5,0) (Sonnet 3.7,33.57) (Sonnet 4.0,33.57) (Gemini 2.5,116.04) (GPT-5,116.04)
    };
    \addplot+[fill=orange!60, draw=orange!80!black] coordinates {
      (Sonnet 3.5,69.0) (Sonnet 3.7,34.7) (Sonnet 4.0,255.87) (Gemini 2.5,156.08) (GPT-5,68.05)
    };
    \legend{Basic prompt, Stressful prompt}
  \end{axis}
\end{tikzpicture}
\caption{L2 task completeness (a) and L3 exploration outcomes (b)}
\label{fig:llm}
\end{figure}
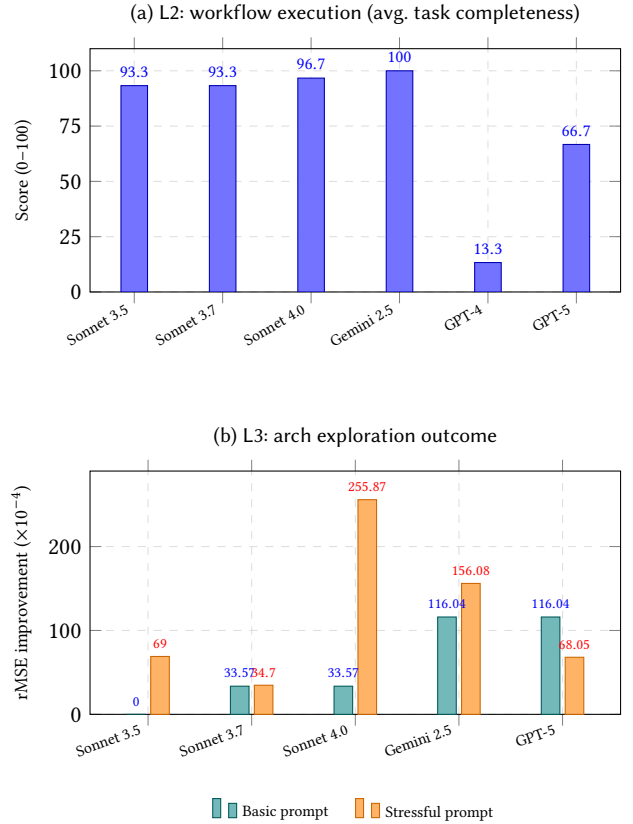

A natural question is how much of A-MLE's behavior comes from the orchestration harness versus the underlying LLM. We hold the agent loop, skills, and prompts fixed and swap the LLM. Figure~\ref{fig:llm} reports L2 task completeness (a) and L3 exploration outcomes (b) across the Claude Sonnet, Gemini, and GPT families.

Three observations stand out. \textit{First}, at the L2 tier, models split cleanly into two clusters based on the capability with high \textit{90s} average task completeness scores. Sonnet $\ge\!3.5$, Gemini~2.5, GPT-5 stood out to be the most capable models for L2 tier. Rest of the other models were unable to operate the workflow loop at all, often hallucinating workflow IDs or failing to wait for asynchronous jobs. \textit{Second}, at the L3 tier, the relationship between model and outcome is more nuanced: Gemini~2.5 and GPT-5 explore most aggressively under the basic prompt, while the Sonnet family tends to play conservatively. \textit{Third}, the effect of prompt stress flips by model family: under stressful, competitive prompts, Sonnet~4.0 surfaces the largest improvement of any configuration ($\sim\!2.6\!\times\!10^{-2}$ rMSE), while GPT-5 becomes \emph{more} conservative and gives back most of its basic-prompt gains.

\subsection{Agent-Surfaced Techniques} \label{subsec:techniques}
Across the model portfolio during the evaluation window, A-MLE surfaced and validated techniques across five families. Table~\ref{tab:techniques} reports counts of distinct models on which each family was attempted and on which the agent's candidate passed human review gating without rework. The agent's strongest results came from \emph{technique transfer} -- recognizing that a technique already validated on one model in the portfolio (e.g. a model arch change or embedding-based feature variant) was likely to transfer to a structurally similar model that had not yet attempted it. The agent struggled on models that had recently undergone non-trivial baseline changes, where its hypotheses were calibrated to a prior version of the model.

\begin{table}[h]
\centering
\small
\begin{tabular}{p{3.8cm}cc}
\toprule
\textbf{Technique family} & \textbf{Attempted} & \textbf{Passed gating} \\
\midrule
Self-supervised pretraining (SSL) & many   & majority \\
Generic optimizer / loss tweaks   & many   & mixed    \\
Embedding-based features          & several & majority \\
Token-mixing architectures        & several & mixed    \\
Architecture scaling              & few    & mixed    \\
\bottomrule
\end{tabular}
\caption{Technique families surfaced by A-MLE. Counts are bucketed: \emph{many} $\geq 8$ models, \emph{several} 3--7, \emph{few} $\leq 2$. ``Passed gating'' counts only candidates that cleared the human review statistical-significance bar without rework.}
\label{tab:techniques}
\end{table}

\subsection{Failure Modes}
We observed five recurring failure modes. 
\begin{itemize}
\item \textbf{Hallucinated APIs:} the agent occasionally invokes plausible but non-existent functions in the training pipeline; pre-flight checks catch these before training launch but consume compute. 
\item  \textbf{Baseline drift:} the agent's offline win is sometimes erased by a concurrent baseline refresh, mitigated but not eliminated by the orchestration harness on rolling-baseline. 
\item  \textbf{Infrastructure fragility:} transient incidents appear to the agent as training divergence, leading to premature abandonment of viable candidates; the retry loop was extended to discriminate these cases. 
\item  \textbf{Over-confident triage:} the agent sometimes promoted a candidate based on a single seed when within-run variance warranted re-running; the auto re-run rule materially reduced this. 
\item \textbf{LLM-specific failures:} weaker base models hallucinate workflow identifiers and pretend to have completed tasks; some models change behavior under stressful prompts in non-monotone ways, as shown in Section~\ref{subsec:llm}.

\end{itemize}
\subsection{Discussion}
The most important design lesson from A-MLE is that the agent's reliability is governed less by the underlying model's reasoning capability and more by the \emph{quality of the surrounding orchestration harness}: the skill library's coverage, the evaluation pipeline's statistical rigor, and the execution layer's resilience to infrastructure noise. We expect this to remain true as base-model capability improves: better LLMs will close the hypothesis-quality gap faster than they close the orchestration harness gap.

%% ---------------------------------------------------------------------
%% Conclusion
%% ---------------------------------------------------------------------
\section{Conclusion and Future Work} \label{sec:conclusion}

A-MLE reframes the bottleneck in industrial ML as the throughput of human iteration rather than the ceiling of any single model, and operationalizes that reframing as a five-stage agent over a shared skill library and execution sandbox. Across our evaluation, A-MLE produced meaningful relative improvements on a majority of models, with the largest gains on the long tail that historically received the least senior attention. 

Three directions stand out for future work. First, we plan to deepen \emph{resilient execution} -- automating more of the discrimination between infrastructure noise and genuine training divergence, and further boosting the productivity and impact from the model iterations. Second, we plan to strengthen \emph{hypothesis generation} via richer domain-specific skills and deeper understanding of the ML model architectures. Third, we plan to extend agentic exploration from a \emph{breadth} regime -- rapidly scaling proven techniques across many models -- into a \emph{depth} regime, in which the agent participates in the design of new architectures and pipelines, with the engineer as architect-in-chief and the agent as implementation and ablation partner.

%% ---------------------------------------------------------------------
%% References
%% ---------------------------------------------------------------------
\bibliographystyle{ACM-Reference-Format}
\bibliography{references}

%% ---------------------------------------------------------------------
%% Appendix
%% ---------------------------------------------------------------------
\appendix
\section{Skill Library Excerpt} \label{appx:skills}
For reference, we list the high-level categories of skills exposed to the agent. Each category contains several typed procedures with structured input/output schemas; we omit the exact signatures.
\begin{itemize}
  \item \textbf{Codebase navigation.} Locate the model architecture implementation, training entry point, identify the active feature group, list recently modified files in the relevant module.
  \item \textbf{Training configuration.} Read the current training config, propose a structured edit, validate the edit against the schema.
  \item \textbf{Launch and monitoring.} Submit a training run with compute estimation, poll status, fetch the most recent logs, distinguish infra errors from training divergence.
  \item \textbf{Evaluation.} Run the offline evaluation suite, compute statistical significance against the rolling baseline, decompose by segment.
  \item \textbf{Proposal authoring.} Assemble a proposal document with experiment design, results, confidence intervals, and a recommendation.
\end{itemize}

\section{Phase Boundary Conventions} \label{appx:phases}
For completeness, we summarize the human-checkpoint conventions at each stage boundary:
\begin{itemize}
  \item \textbf{After hypothesis generation.} Reviewer approves a hypothesis or substitutes one of their own.
  \item \textbf{After exploration strategy.} Reviewer confirms the planned experiment set and compute allocation.
  \item \textbf{After experiment execution.} Reviewer confirms the run results and statistical analysis before proposal authoring.
  \item \textbf{After proposal authoring.} Reviewer revises the proposal as needed and routes it through the standard review process.
\end{itemize}

\end{document}